\documentclass[10pt, review]{elsarticle}

\usepackage{microtype}
\usepackage{graphicx}
\usepackage{booktabs} 
\usepackage{amssymb}
\usepackage{amsmath}
\usepackage{subcaption}
\usepackage{graphicx}
\usepackage{soul,color}
\usepackage{multirow}
\usepackage{bm} 
\usepackage{balance}
\usepackage{comment}
\usepackage{hyperref}

\newcommand{\addReviewer}[2]{
  \expandafter\newcommand\csname #1\endcsname[1]{{\bf \color{#2} \expandafter\MakeUppercase #1:\,##1}}
}

\definecolor{asparagus}{rgb}{0.53, 0.66, 0.42}
\definecolor{alizarin}{rgb}{0.82, 0.1, 0.26}
\addReviewer{jary}{asparagus}
\addReviewer{simone}{alizarin}

\newif\ifproofread

\newcommand{\changemarker}[1]{%
\ifproofread
\textcolor{blue}{#1}%
\else
#1%
\fi
}

\begin{document}
\proofreadfalse

\begin{frontmatter}
\title{Continual Learning with Invertible Generative Models}

\author[sapienza]{Jary Pomponi \corref{cor1}}
\cortext[cor1]{Corresponding author: jary.pomponi@uniroma1.it}

\author[sapienza]{Simone Scardapane}

\author[sapienza]{Aurelio Uncini}

\address[sapienza]{Department of Information Engineering, Electronics and Telecommunications (DIET), Sapienza University of Rome, Italy}

\begin{abstract}
    Catastrophic forgetting (CF) happens whenever a neural network overwrites past knowledge while being trained on new tasks. 
    Common techniques to handle CF include regularization of the weights (using, e.g., their importance on past tasks), and rehearsal strategies, where the network is constantly re-trained on past data. Generative models have also been applied for the latter, in order to have endless sources of data. In this paper, we propose a novel method that combines the strengths of regularization and generative-based rehearsal approaches. Our generative model consists of a normalizing flow (NF), a probabilistic and invertible neural network, trained on the internal embeddings of the network. By keeping a single NF throughout the training process, we show that our memory overhead remains constant. In addition, exploiting the invertibility of the NF, we propose a simple approach to regularize the network's embeddings with respect to past tasks. We show that our method performs favorably with respect to state-of-the-art approaches in the literature, with bounded computational power and memory overheads.
\end{abstract}


\end{frontmatter}

\section{Introduction}
\label{section:introduction}


One of the major open problems in deep learning is the so-called catastrophic forgetting (CF) \cite{MCCLOSKEY1989109, Ratcliff1990ConnectionistMO, french1999catastrophic}. It is the tendency of a neural network (NN) to forget past learned information when training on new tasks. This problem is intrinsically connected with the continual learning (CL) property of a NN, which is the ability of a NN to learn consecutive information without forgetting previously stored knowledge.    


Overcoming, or mitigating, CF is a key step in order to achieve a more general artificial intelligence; a system should be able to learn a sequence of tasks and remember them, following the lifelong learning paradigm \cite{THRUN199525}. This is a key problem because real world tasks continually evolve and, often, it is not possible to train a NN from scratch. Without efficient methods to overcome CF, online training in a lifelong learning scenario is not possible. Recently, there has been a resurgence of interest in this area due to its importance.

One of the first attempts to mitigate CF consisted in storing past examples and replaying them into the model while learning new information \cite{robins1995catastrophic}. These \textit{rehearsal} methods have been improved over the years, with more complex memory systems and hybrid approaches emerging. In particular, many \textit{pseudo-rehearsal} methods have been proposed, in which the external memory is replaced with a generative model, capable of generating endless samples from the past. Pseudo-rehearsal algorithms, however, require generative models that are both easy to train (in order to provide high-quality data) and simple to condition on the task (in order to avoid having one generative model for each task). Both of these conditions are extremely challenging in practice.

To overcome (or complement) these limitations, many \textit{regularization} methods have been studied over the last years, in which additional loss terms are used to mitigate CF. In elastic weight consolidation (EWC, \cite{Kirkpatrick2017overcoming}), for example, past weights are used to regularize the training process, by slowing down the modification of important weights, where the importance is quantified based on a Fisher information criterion. Alternative regularization strategies are achieved by acting on the previous outputs \cite{li2017learning} or gradients \cite{lopez2017gradient}. Recently, it was also shown that state-of-the-art performance can be achieved by acting on the internal embeddings (i.e., the activations before the classification layer) \cite{pomponi2020er}. A more complete overview of CL methods is provided later on in Section \ref{section:methods}, or in \cite{parisi2019continual}.


In this paper, we propose a novel CL method aiming to combine the benefits of pseudo-rehearsal and regularization strategies. Our algorithm can be summarized in three points:

\begin{enumerate}
    \item Similarly to pseudo-rehearsal, we store information on past tasks by training an auxiliary generative model. Instead of training it on the input space, however, we train it to generate samples from the internal embeddings of the network, simultaneously with the main classifier.
    \item We use a normalizing flow (NF, \cite{papamakarios2019normalizing}) as generative model. NFs are invertible networks that can perform sampling and density estimation in both directions. In this way, the NF can be trained efficiently, with no need for additional components such as in generative adversarial networks.
    \item Finally, we use the sampled embeddings from the trained NF to perform regularization with respect to the past tasks (see Fig. \ref{fig:regularization} later on).
\end{enumerate}

We claim that (i) training the generative model in the embeddings' space is significantly easier (both in its design and in its optimization) compared to the input space, and (ii) that regularization makes better use of past information stored in the generative model, compared to simply augmenting the mini-batch with new data (similarly to \cite{pomponi2020er}). In our experimental evaluation, we validate these two claims, and we show that our method performs favourably (or better) than several state-of-the-art approaches, while requiring significantly less memory and computational overhead. We make our code publicly-available.\footnote{\url{https://github.com/jaryP/Continual-Learning-with-Invertible-Generative-Models}}
\section{Related Works}
\label{section:methods}

The methods for overcoming CF can be categorized, in line with \cite{parisi2019continual} and \cite{maltoni2019continuous}, in three broad groups. We underline that the boundaries are not always defined, with many methods, including ours, exploiting two or more of these strategies.

\begin{itemize}
    \item \textbf{Architectural Strategies}: methods that use specific architectures, layers, activation functions and/or weights freezing/pruning, and eventually grow the architecture when needed (e.g., \cite{rusu2016progressive, pomponi2021structured}). For example, Hard Attention on Task (HAT, \cite{serra2018overcoming}) uses an attention mechanism in order to route the information/gradient flow in the NN and preserve the weights associated to past tasks.
    \item \textbf{Rehearsal strategies}: in this case, past examples are stored and later replayed in the current mini-batches to consolidate the network. In order to avoid having to explicitly store past examples, which requires a growing memory, \textit{pseudo-rehearsal} algorithms \cite{robins1995catastrophic} craft them on the fly, most notably exploiting generative models \cite{shin2017continual}. \changemarker{Rehearsal approaches aren't just the ones that use saved images to replay them while training, but any approach that uses saved images to fight CF; e.g. the saved images can also be used to regularize the internal state of the model, as proposed in \cite{pomponi2020er} or \cite{zhu2021prototype}.}
    \item \textbf{Regularization techniques}: in this case, popularized by elastic weight consolidation (EWC, \cite{Kirkpatrick2017overcoming}), the loss function on the current task is extended with a regularization penalty to selectively consolidate past information or slow the training on new tasks. Broadly speaking, regularization methods are easy to implement, but they require carefully selecting what information is regularized, and how.
\end{itemize}

The method we propose in this paper is at the boundary of pseudo-rehearsal and regularization strategies, so we focus on these two classes below. 

Learning Without Forgetting (LWF, \cite{li2017learning}) is one of the earliest regularization methods. It attempts to alleviate CF by stabilizing the output layer using knowledge distillation. Other well-known regularization methods are EWC, which applies a soft structural regularization computed between the weights' importance relative to the past tasks and the current weights, and Synaptic Intelligence (SI, \cite{zenke2017continual}), a modification of EWC, which uses the difference between the current weights and their trajectory calculated during the training. Other methods include Gradient Episodic Memory (GEM, \cite{lopez2017gradient}), Averaged-GEM \cite{chaudhry2018efficient}, and the recently proposed Embedding Regularization (ER, \cite{pomponi2020er}). 

In GEM, the external memory is populated with past examples that are used to regularize the direction of the current gradients, in order to move the weights in a region of the space in which all the tasks are satisfied. This method is capable of improving past scores, but it requires solving a complex minimization problem at every step, which does not scale well with the number of tasks. ER is a regularization technique in which the external memory contains past examples and their associated embeddings, extracted at the end of the training process on the associated task. The memory is used to impose a penalty to constrain the current embeddings to lie in the vicinity of the past ones. This method is extremely fast and requires little memory. The approach we propose in this paper follows the philosophy of ER to act at the level of the embeddings, instead of single weights or outputs. Another novel and interesting approach is proposed in \cite{ebrahimi2020adversarial}, in which the authors use an Adversarial Continual Learning (ACL) approach: it aims to alleviate CF by learning a disjoint latent space representation composed of a task-specific latent space for each task and a task-invariant feature space for all tasks.

A more challenging set of methods are pseudo-rehearsal ones. In \cite{shin2017continual}, the authors proposed a method which consists of two modules: a deep generative model and a task solver. In this way, samples from past tasks can be generated using the generative model and interleaved with information from the new tasks; the solver is used to predict the label associated to the generated images in order to regularize the network. In \cite{kang2020discriminative} a similar approach, but based on a Variational Autoencoder (VAE), is proposed: it consists of a VAE and an external NN, which learns to replicate the distribution of the embeddings associated to a task; this external NN can be used to generate images associated to past tasks and reduce CF. \changemarker{Pseudo-rehearsal methods based on VAE are evaluated mostly on datasets of relatively low complexity, due to the limit of those generative models to create photo-realistic images, but, on the other hand, are easier to train if compared to GANs \cite{zhao2017infovae, zhao2017towards}}. Whether these generative approaches can scale up to more complex domains is still an open problem. 

The method we propose here can be considered a pseudo-rehearsal one, but we focus on a more recent class of generative models \cite{kingma2018glow}, and we apply them at the level of embeddings instead of in the input space; \changemarker{this should give us the ability to create realistic-samples while having an easy to train generative model}. We note that in the literature on generative models, a number of authors have considered similar combinations of autoencoders with a generative model on their latent space. In \cite{rezende2015variational}, the authors applied a NF to learn a VAE prior. This idea has been further studied in \cite{kingma2016improved}, where the authors proposed a new NF which scales well for high-dimensional embedding spaces. Similar to our proposal, in \cite{guo2019auto} the authors proposed a model that uses an adversarial generative model in the embedding space to generate high-resolution images.

A relatively new and emerging area of study researches how to mitigate CF when the tasks' boundaries are not known. In this area we highlight \cite{aljundi2019task}, in which the authors proposed a task free approach to continual learning, using a regularization-based memory. In \cite{zeno2018task}, a task agnostic Bayesian method was proposed, demonstrating the ability of probabilistic models to handle ambiguous task boundaries. Finally, \cite{rao2019continual} has introduced the novel idea of unsupervised learning in a lifelong scenario.


Many other methods exist; for a complete review of existing methods see \cite{parisi2019continual, masana2022class, de2021continual}. 

    \begin{table}[!t]
    \caption{Basic notation used in the paper}
    \label{tab:symbols}
    \centering
    \resizebox{0.8\linewidth}{!}{%
    \begin{tabular}{cc}
    \hline
    $\mathbf{x}$ & Input image \\ 
    $y^t$ & Label w.r.t. the task $t$  \\ 
    $y^d$ & Label w.r.t. all tasks  \\ 
    $S_t(\cdot)$ & Classification head for task $t$ \\
    $E(\cdot)$ & Encoder  \\
    $E_c(\cdot)$ & \shortstack{Classification encoder, defined as $E_c = f_c \circ E$} \\  
    $E_r(\cdot)$ & Reconstruction encoder, defined as $E_r = f_r \circ E$ \\  
    $D(\cdot)$ & Decoder \\  
    $p_u(u)$ & Prior for the NF
    \end{tabular}%
    }
    \end{table}

\section{Proposed method}
\label{section:proposed}

In this section, we describe our proposed PRER method. Section \ref{sec:motivation} provides some motivation for the method. Section \ref{section:formulation} describes the general CL setting we consider. Section \ref{section:components} overviews our method and the way we perform the training. Then, Sections \ref{sec:regularization} and \ref{section:nf} go in-depth into the regularization and generative components, respectively. To help reading, we summarize some common notation in Table \ref{tab:symbols}.

\subsection{Motivation}
\label{sec:motivation}
In a sense, rehearsal methods are close to optimal, because in the limit of a very large memory they recover a standard \changemarker{classification approach, in which the model is trained on the whole dataset}. On the other hand, these methods require a memory that, usually, grows linearly with the number of tasks. 

Pseudo-rehearsal methods try to overcome these limitations by substituting the memory with a generative model and doing parameter sharing on the generative model, which is incrementally trained on all the tasks and constrained to remember the information about the tasks encountered so far. However, this creates a new set of challenges: (i) the CF problem is removed from the NN, but the generative model itself potentially suffers CF; (ii) doing parameter sharing on real-world images can be difficult. 

The aim of this paper is to propose a generative approach which does not work directly on the input, and that can be used to regularize the model instead of simply augmenting the dataset. The key idea of the proposed Pseudo-Rehearsal Embedding Regularization (PRER) is to use a generative model to sample new embeddings associated to past tasks, and to use a decoder to reconstruct the associated images. To regularize the network, we (i) reconstruct the images associated to the embeddings, (ii) calculate the embeddings given by the current network; (iii) force the old and the new embeddings to be as close as possible (by moving the current one in the direction of the past ones). In our experimental section, we show that the generative model is not prone to CF, while the encoder is easily regularized using the above-mentioned process.


\begin{figure}[t]
\vskip 0.2in
\begin{center}
\centerline{\includegraphics[width=0.95\columnwidth]{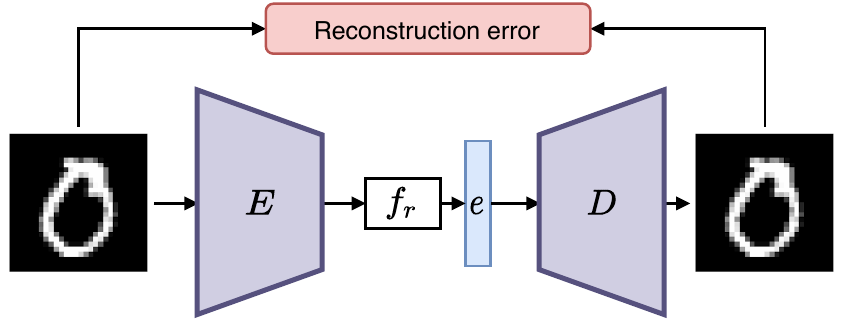}}
\caption{The autoencoder training process. The images   are projected into a subspace called embedding space and reconstructed to match as much as possible the input images.}
\label{fig:train_ae}
\end{center}
\vskip -0.2in
\end{figure}
\begin{figure}[t]
\vskip 0.2in
\begin{center}
\centerline{\includegraphics[width=0.95\columnwidth]{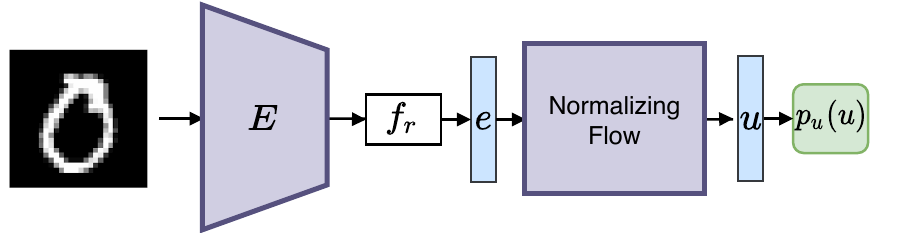}}
\caption{The NF training process. The generative model is trained to transform the embeddings into a vector which contains only values that match the base distribution $p_u(u)$, called prior.}
\label{fig:train_nf}
\end{center}
\vskip -0.2in
\end{figure}

\subsection{Problem formulation}
\label{section:formulation}
The following formulation of the CL scenario is similar to the one proposed in \cite{lopez2017gradient}. We receive a sequence of tasks $t=1, \dots, M$, each one composed by a set of triples $\{(\mathbf{x}_i, t, y^t_i)\}_{i=1}^S \in \mathcal{X} \times \mathbb{N^+} \times \mathcal{Y}$, where $\mathbf{x}_i$ is a sample, $t$ is an integer identifying the current task, and $y^t_i$ is the label of $\mathbf{x}_i$ w.r.t. the current task $t$. Within this formulation, the tasks never intersect, and a new task is collected only when the current one is over. The input and the labels can belong to any domain, although most benchmarks in the CL literature have considered the image domain in a classification setting \cite{parisi2019continual}.

In this paper we focus on a CL scenario in which each task has its own classifier, meaning that only a portion of the NN, that we call the encoder $E$, is shared.
There is no class overlap among different tasks, and accuracy is computed separately for each one. This is done by creating a leaf branch for each task, called head, which classifies images associated to that task.
 Mathematically, we have a model $f(\mathbf{x}, t) = S_t(E(\mathbf{x}))$, where $\mathbf{z} = E(\mathbf{x})$ is the encoder and $S_t(\mathbf{z})$ a task-specific classifier. We call $\mathbf{z}$ the \textit{embedding} of $\mathbf{x}$.
In a naive setting, when we receive a new task $t$, we minimize a task-specific loss starting from the current encoder $E$ and a randomly-initialized $S_t$:
\begin{equation}
\mathcal{L}_c^t = \frac{1}{S}\sum_{i=1}^S 
L(f(\mathbf{x}_i, t), y^t_i) \,,
\label{eq:classifier_loss}
\end{equation}
where $L$ is a suitable loss (e.g., cross-entropy). CF appears whenever training on the current task $t$ degrades the performances on previous tasks $1, \ldots, t {-} 1$.


\subsection{Components of PRER}
\label{section:components}
For PRER, we augment the encoder $E$ and classifier heads described in Section \ref{section:formulation} with two additional modules:
\begin{enumerate}
    \item \textbf{Decoder}: a decoder $D$ is trained to approximately invert the encoder $E$  (i.e., $D(E(\mathbf{x})) \approx \mathbf{x}$), by minimizing the mean squared error (MSE) between the original images and the embeddings. We refer to the combination of encoder and decoder as the \textit{autoencoder} part of the model (see Fig. \ref{fig:train_ae}).
    \item \textbf{Generative model}: a generative model, implemented as an NF (formally described in Section \ref{section:nf}), is trained to approximate the distribution $p_t(\mathbf{z})$
    of embeddings up to the current task $t$. 
    
\end{enumerate}

Empirically, we found that sharing the entire encoder $E$ for the classification heads and the decoder results in slightly reduced performance. Inspired by recent works on contrastive learning \cite{grill2020bootstrap}, we solve this by projecting the output of $E$ using two  separate \textit{projection heads} $f_c$ and $f_r$, resulting in a classification encoder $E_c = f_c \circ E$, and a reconstruction encoder $E_r = f_r \circ E$, where only the \textit{backbone} encoder $E$ is shared. The classifier for task $t$ is then defined as $S_t \circ E_c$ (for a task-dependent $S_t$), while the \changemarker{autoencoder} is defined as $D \circ E_r$.

Given a new task, the training process is divided into three phases described below. We found that splitting the training into three separate processes (instead of joint training) helps the stability and improves the results since each block works with the others already optimized. 

\noindent \textbf{1) Classifier training}: In the first stage, the current task-specific head $S_t$, \changemarker{as well as the encoder $E_c$,} are trained to solve the current task $t$ by minimizing a modified version of Eq. \eqref{eq:classifier_loss}, in which the encoder is the one which uses the projection head associated to the classification:

\begin{equation}
\mathcal{L}_c^t = \frac{1}{S}\sum_{i=1}^S 
L(S_t(E_c(\mathbf{x}_i)), y^t_i) \,.
\label{eq:classifier_loss_1}
\end{equation}
\\
\textbf{2) Autoencoder training}: When the training process of the classifier head is over, we train the decoder $D$ and the projection function $f_r$ to minimize the reconstruction error on the current task. In this step, we keep the encoder fixed to minimize CF. As loss function we use the MSE between the real images and the reconstructed ones:
\begin{equation}
    \mathcal{L}^t_{\text{ae}}(\mathbf{x}) = \frac{1}{S}\sum_{i=1}^S \lVert \mathbf{x}_i - D(E_r(\mathbf{x}_i))) \rVert^2 
    \label{eq:autoencoder_loss}
\end{equation}
\noindent The training process of the autoencoder is visually shown in Fig. \ref{fig:train_ae}. The decoder function can also be conditioned to guide the reconstruction of the embeddings. Denote by $y^d$ a label defined on the union of all possible classes (e.g., if task 1 has 2 classes and task $2$ has 3 classes, $y^d$ is defined over $5$ possible classes). We guide the decoder by providing $y^d$ as input together with $\mathbf{x}$, i.e., $\widetilde{\mathbf{x}} = D(E_r(\mathbf{x}, y^d))$.
\\



\noindent \textbf{3) Generative model training}: In the last stage we train the generative model to learn the underlying distribution of the embeddings generated by $E_r(\cdot)$. The generative model and the training procedure are described in detail in Section \ref{section:nf}, and shown in Fig. \ref{fig:train_nf}. We underline that we use a single generative model, so that the memory footprint is constant and the model is capable of generating embeddings from all the tasks $\leq t$, which will be reconstructed by the decoder trained in the previous step.
After training the generative model on task $t$ we have a distribution $p_t(\mathbf{z})$, which is capable of generating embeddings from each task encountered so far. The generative model can also be conditioned, to force the sampling of an embedding associated to a specific class $y^d$. 

\subsection{Regularization}
\label{sec:regularization}
When training on a task which is not the first one, we need to regularize the network to avoid CF. In the proposed method, we use the NF to sample embeddings from the previous tasks, which we use in two ways: (i) we augment the dataset while training the decoder and the NF (steps 2 and 3 before); (ii) we add a regularization term to force the encoder  to maintain these embeddings while training on the new task, similarly to \cite{pomponi2020er}, in step 1.

To do so, at the beginning of a new task $t > 1$, we sample a set of images, by combining the generative model and the decoder. We denote this set of $N$ tuples as $\mathcal{M}_{<t}$: 

\begin{equation}
    \mathcal{M}_{<t} = \{ ( \hat{\mathbf{x}}_i, \hat{\mathbf{z}}_i) \}_{i=1}^{N} \,.
    \label{eq:ae_reg}
\end{equation}

\begin{figure}[t]
\vskip 0.2in
\includegraphics[width=\columnwidth]{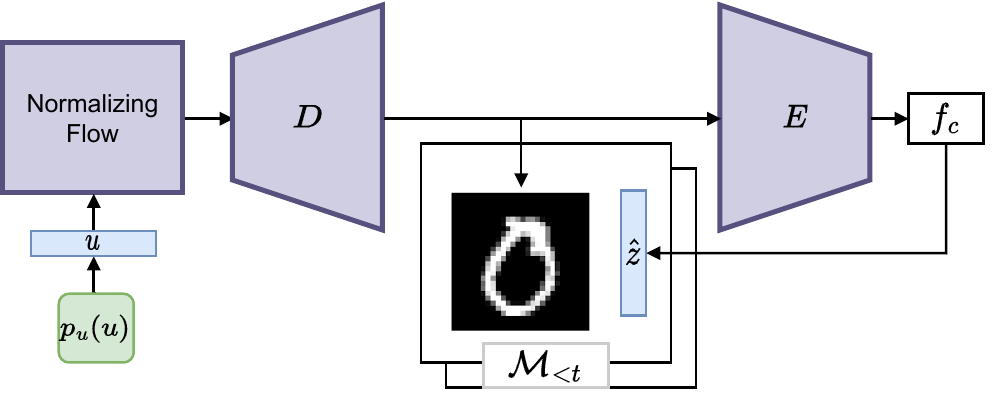}
\caption{The image shows how the external memory is generated. In the first step, the NF takes a vector drawn from the prior $p_u(u)$ and generates the associated embedding vector, used to reconstruct the image $\mathbf{x}$. Then, the image is added to the memory, along with the features extracted from $E_c(\mathbf{x})$.}
\label{fig:memory_creation}
\vskip -0.2in
\end{figure}

\noindent For each tuple $i$, we first sample an embedding $\mathbf{z}_i \sim p_{t-1}(\mathbf{z})$ from the NF. We then store in $\mathcal{M}_{<t}$ the reconstructed image $\hat{\mathbf{x}}_i = D(\mathbf{z}_i)$ and the current embedding $\hat{\mathbf{z}}_i = E_c(\hat{\mathbf{x}}_i)$. Once we have generated the synthetic dataset $\mathcal{M}_{<t}$, we use it to regularize the training as described below. This sampling process is visually shown in Fig. \ref{fig:memory_creation}. Note that it is not necessary to generate an entire set of synthetic images $\mathcal{M}_{<t}$ prior to the training, because the synthetic images can be generated on-the-fly. However, we found no practical benefit in doing so, and pre-storing an entire set is more computationally efficient. 




\begin{figure}[t]
\vskip 0.2in\centering
\includegraphics[width=0.8\columnwidth]{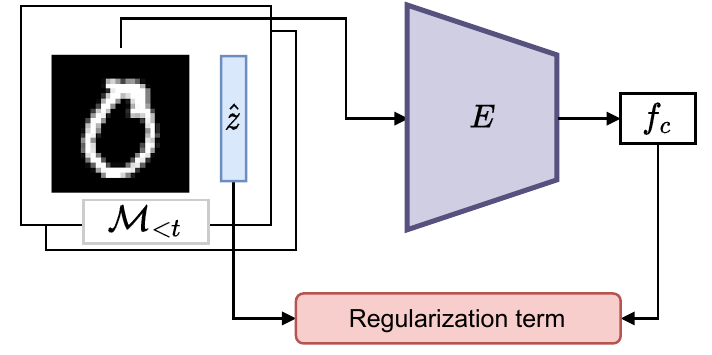}
\caption{Here, the regularization process is displayed. The reconstructed images are used to calculate the current embedding vectors, and the results are used to calculate the regularization term. For further explanations see Section \ref{sec:regularization}.}
\label{fig:regularization}
\vskip -0.2in
\end{figure}

While training the autoencoder and the NF on a new task, we overwrite a portion of each mini-batch with some past samples from the memory. In this way, the decoder and the NF do not suffer CF, because they are trained each time on all the tasks seen so far. Regarding the regularization of the classifier, we apply the same technique proposed in ER \cite{pomponi2020er}. Consider a generic image $\mathbf{x}$ and its embedding $\mathbf{z}$ produced by $E_c(\mathbf{x})$,
the regularization term is calculated as:
\begin{equation}
    R({\mathbf{z}}, \mathbf{x}) = d(\mathbf{z}, E_c(\mathbf{x}))\,,
    \label{eq:embedding_regulatization}
\end{equation}
where $d$ is a suitable distance function. We augment the classification loss in \eqref{eq:classifier_loss_1} with the regularization term as:

\begin{equation}
\begin{aligned}
    \widetilde{\mathcal{L}}_c^t &=
    \frac{1}{S} \sum_{i=1}^S L(S_t(E_c(\mathbf{x}_i)), y_i^t)  \\
    & + \frac{\beta}{\vert\mathcal{M}_s\vert}
    \sum_{(\hat{\mathbf{x}}, \hat{\mathbf{z}}) \in \mathcal{M}_{<t}}
    d(\hat{\mathbf{z}}, E_c(\hat{\mathbf{x}}))\,.
\end{aligned}
    \label{eq:autoencoder_loss_er}
\end{equation}
\noindent where $\beta$ is a parameter that regulates the importance of the regularization term. This ensures that the embeddings used for classifying the images do not change during the training. This step is applied only while training the classifier, and it is summarized in Fig. \ref{fig:regularization}. Practically, at each iteration of training we sample a mini-batch from the dataset and an independent mini-batch from the memory.

 

\subsection{Normalizing Flows}
\label{section:nf}

To complete the specification of our PRER method, we need to describe a specific generative model to estimate samples from the embedding space. While most generative methods could be used here, we have found NFs to be particularly effective for the task. Introduced in \cite{Tabak2010Density} and \cite{Tabak2013Nonparametric}, and popularized by \cite{rezende2015variational} and \cite{dinh2014nice}, NFs are probabilistic models describing the transformation of a probability distribution into a more complex one using a sequence of differentiable, invertible mapping functions. Depending on the type, a NF is capable of efficiently performing sampling or density estimation in either direction \cite{papamakarios2019normalizing}.

Let $\mathbf{u} \sim p_{u}(\mathbf{u})$ be a real $d$-dimensional distribution which is easy to sample from, and $T$ a transformation $T : \mathbb{R}^d \rightarrow \mathbb{R}^d$. The key requirements for $T$ to be a NF are: 1) $T$ must be invertible, with an inverse denoted by $T^{-1}$, and 2) both $T$ and $T^{-1}$ must be differentiable. We can obtain samples from $p_z(\mathbf{z})$ by drawing samples from the easier $p_u(\mathbf{u})$ and then computing $T(\mathbf{u})$ (forward mapping). Alternatively, we can `normalize' a known sample $\mathbf{z} \sim p_z(\mathbf{z})$ by applying the inverse transformation $T^{-1}(\mathbf{z})$ (inverse mapping).\footnote{As mentioned in \cite{papamakarios2019normalizing}, the terms ``forward'' and ``inverse'' are simply a convention.} Additionally, we can evaluate the likelihood of a known sample $\mathbf{z}$ as:
\begin{align}
    p_{z}(\mathbf{z}) = p_{u}(\mathbf{u}) \Big\vert \text{det}\ J_{T}(T^{-1}(\mathbf{x})) \Big\vert ^ {-1} \,,
    \label{eq:change_of_variable}
\end{align}
where $J_{T^{-1}} \in \mathbb{R}^{d \times d} $ is the Jacobian matrix of all partial derivatives of $T^{-1}$. The prior distribution $p_{u}(\mathbf{u})$ is generally chosen as an Isotropic Normal distribution. $T$ is instead obtained by composing multiple simpler, invertible transformations $T_1$, $\ldots$, $T_L$, resulting in $T = T_L \circ T_{L-1} \circ \dots \circ T_1$. The arbitrarily complex density $\mathbf{z}$ can be constructed from a prior distribution by composing several simple maps and then applying Eq. \eqref{eq:change_of_variable}.



As stated before, a NF can perform both sampling and density estimation in both directions. Depending on the specific NF, not all of these operations are necessarily easy to compute \cite{papamakarios2019normalizing}. This is due to the fact that to achieve them simultaneously, the model needs both a simple forward mapping, a simple inverse mapping, and the Jacobians have to be easy to compute. In general, this is not always possible, mostly due to a computational trade-off. Hence, a NF model needs to be built depending on the application.

In our case, we need efficient density estimation and sampling in the inverse mapping (for training), but only efficient sampling in the forward direction (see Fig. \ref{fig:train_nf} and \ref{fig:regularization}). We use a setup similar to \cite{dinh2016density,kingma2018glow}, whose layers we briefly summarize below. Denote by $\mathbf{u}_{j+1} = T_j(\mathbf{u}_j)$ the generic $j$th block of the NF. We build the overall NF by interleaving three types of invertible transformations.

\noindent\textbf{1) Coupling Layer and Affine transformation}: Proposed in \cite{dinh2014nice, dinh2016density}, a coupling layer consists in a powerful reversible transformation, where the forward and the inverse mappings are computationally efficient. Here, we use the version presented in \cite{kingma2018glow}. Consider a generic split of the input vector $\mathbf{u}_j = \begin{bmatrix}\mathbf{a} \\ \mathbf{b} \end{bmatrix}$. A coupling layer is defined as: 
\begin{align*}
    & \mathbf{t} = f_{\theta}^t(\mathbf{a}) \\
    & \log \mathbf{s} = f_{\theta}^s(\mathbf{a}) \\
    & \mathbf{c_b} = \text{exp}(\log \mathbf{s})\odot \mathbf{b} + \mathbf{t} \\
    & \mathbf{u}_{j+1} = \text{concat}(\mathbf{a}, \mathbf{c_b})
\end{align*}
\noindent where $f^t_\theta (\cdot)$ and $f^s_\theta (\cdot)$ are generic transformations implemented via a NN, and $\odot$ is the element-wise multiplication. The $\log \text{det}$ of a coupling layer is simply $\sum_i \log s_i$. The main advantage of a coupling layer is that the functions $f^t_\theta (\cdot)$ and $f^s_\theta (\cdot)$ do not have to be invertible. The main advantage is also a disadvantage, since, due to the simplicity of the affine transformation, a NF implemented with this technique needs multiple layers and blocks in order to have enough expressive power to transform any input into $p_u(\mathbf{u})$. In our case the NF is applied on the embedding space, and we will see that the required number of layers necessary to achieve good results is small.  

\noindent\textbf{2) Random Permutation}: since only a portion of the input is modified in each block, it is required to randomly permute the output of a block in order to modify a new set of parameters in the following block, i.e., $T_j(\mathbf{u}) = \mathbf{P}\mathbf{u}$, where $\mathbf{P}$ is a fixed permutation matrix. Clearly, $T^{-1}_j(\mathbf{u}) = \mathbf{P}^T\mathbf{u}$, and the transformation has a unitary determinant.

\noindent\textbf{3) Invertible Batch Norm}: As in \cite{dinh2016density}, we also apply batch normalization, but on the input of the coupling layer instead of the output. It acts as a linear re-scaling of each parameter, thus it can be easily inverted and included in the Jacobian computation. The scaling is done in the following way: 
\begin{align*}
 \mathbf{u}_{j+1} = \frac{\mathbf{u} - \bm{\mu}}{(\sqrt{\bm{\sigma}^2 + \epsilon)}}
\end{align*}
and the parameters are iteratively estimated as: 
\begin{align*}
    & \bm{\mu} = m \bm{\mu} + (1-m)\bm{\mu}_b \\
    & \bm{\sigma} = m \bm{\sigma} + (1-m)\bm{\sigma}_b 
\end{align*}
\noindent where $m \in [0, 1]$ is the momentum, while $\bm{\mu}_b$ and $\bm{\sigma}_b$ are, respectively, the mean and the standard deviation of the current mini-batch. On the first batch, the parameters are initialized as $\bm{\mu} = \bm{\mu}_b$ and $\bm{\sigma}=\bm{\sigma}_b$, and they are updated with each new batch during the training. The $\log \text{det}$ is computed as: 
\begin{align*}
    \Big( \prod (\bm{\sigma}^2 + \epsilon) \Big)^{-0.5}
\end{align*}
\noindent with $\epsilon > 0$ a parameter to avoid zero multiplication. 

Note that, in our case, the NF can be conditioned on $y^d$. To implement this, we condition the first coupling layer by passing the class $y^d$ as an additional argument to $f^t_\theta (\cdot)$ and $f^s_\theta (\cdot)$. The idea is similar to what is done in \cite{winkler2019learning}. Practically, the conditioning is done by building a one-hot vector with respect to the maximum number of separate classes we presume to observe. 

Furthermore, we implement a multi-scale architecture as explained in \cite{dinh2016density, kingma2018glow}. The NF is partitioned in several levels, each one composed by multiple blocks of permutation, coupling layer and batch norm, and at the end of each level the output is split into two equal chunks: the parameters in the first chunk are sent directly to the output, while the other chunk flows into the next level for further processing (if another level is present). This architecture improves the convergence and the stability, and it results in a smaller number of parameters.


Many other ways of building a NF exist; for an in-depth review of the NF literature we refer to \cite{papamakarios2019normalizing} and \cite{Kobyzev2020NormalizingFA}. 


\section{Experiments}

\subsection{Datasets and metrics}

To evaluate the proposed method, we consider three different datasets: MNIST, SVHN \cite{Netzer2011}, and \changemarker{CIFAR10-100} \cite{Krizhevsky09learningmultiple}. We evaluate the methods under the previously described multi-head CL scenario; to do so, being $C$ the classes of a dataset, these are grouped in $M$ sets, each one making up a task containing $c_m \in \mathbb{N}_+$ classes, with $c_m > 1$, giving: $M = \lceil\frac{C}{c_m}\rceil$ (
with this formulation each task contains the same number of classes, with the exception of the last one if $C\ \text{mod}\ c_m \ne 0$). 
In particular, we split the labels by grouping the original ones in an incremental way. 
\changemarker{For all the datasets, with the exception of CIFAR100, we set $c_m = 2$, generating 5 tasks. Regarding CIFAR100, we set $c_m=10$, creating $10$ tasks.}

To evaluate the efficiency and to compare the methods two metrics from \cite{diaz2018don} have been used. All these metrics are calculated on a matrix $R \in \mathbb{R}^{M \times M}$, where $M$ is the number of tasks, and each entry $R_{ij}$ is the test accuracy on task $j$ when the training on task $i$ is completed. The chosen metrics are summarized below.

\textbf{Accuracy}: it is the average accuracy obtained on the test set of each task, after the training on the last task is over, and it is calculated as: 
\begin{align*}
 \text{Accuracy} = \frac{1}{M}\sum_{j=1}^M R_{Mj} \,.
\end{align*}
\noindent This metric aims to show the final average accuracy obtained by the final model, and does not takes into consideration the evolution of the scores.

\textbf{Backward Transfer (BWT)}: it measures how much information from the old tasks is remembered during the training on a newer one. It is calculated as:
\begin{align*}
 \text{BWT} = \frac{\sum_{i=2}^{M} \sum_{j=1}^{i-1} (R_{ij} - R_{jj})}{\frac{1}{2} M (M-1)} \,.
\end{align*}
This metric can be greater than zero, meaning that not only the model is remembering everything about past tasks, but it also improves past scores. In our scenario, this phenomenon is rare, since heads associated with past tasks are no longer trained. This metric is also calculated when the training on all the tasks is over.

We use these metrics because they embed all the important aspects of the CL problem: the ability of an approach to mitigate CF and to classify correctly the past tasks, but also the capability to train the model on the current one. They are both important because a model with high accuracy and low BWT is a model not capable of alleviating CF; on the other hand, low accuracy and high BWT tells us that the constraints applied to the model are too restrictive, blocking the training of the current task.

\subsection{Baselines}
We compare our method to the Naive approach --training on all the tasks sequentially without mitigating the CF problem-- and to three established or related rehearsal and regularization approaches: \changemarker{Replay \cite{chaudhry2019tiny}}, EWC \cite{Kirkpatrick2017overcoming}, GEM \cite{lopez2017gradient}, GFR-IL \cite{liu2020generative},  ER \cite{pomponi2020er}, \changemarker{and Supermask in Superposition (SupSup) \cite{wortsman2020supermasks}}. \changemarker{As an additional baseline, we used our proposal to generate images to regularize the model using the replay approach; this method is called PRER-R. Regarding SupSup, being an architectural approach that freezes the weights of the model, it has no BWT.} 

Concerning ER, we used a slightly modified version of the original algorithm to align it to our scenario. In the original paper, the memory is filled with past images, and during the regularization step only a subset of these images is used to calculate the regularization parameter; in this paper, to align ER to the other rehearsal methods, all the images saved in the external memory are used. This is a small modification that improves slightly the result presented in the original paper. 

For each method, we searched for the best hyper-parameters, which are exposed in Section \ref{section:Hyperparameters}.
\subsection{Models and training}

We trained all the models using Adam as optimizer \cite{kingma2014adam}, with the learning rate equal to $0.001$. To avoid injecting some past information into the training of the current task, the parameters of the optimizer are restored after each task. We also trained the autoencoder and the NF of PRER with the same setting. 
The training process for the encoder $E_c(\cdot)$ and the heads proceeds for a fixed number of epochs, and at the end of the training, the best model is saved and used for future tasks. For PRER we train $E_r(\cdot)$ and the NF until the respective losses stop improving for 5 consecutive epochs.

For MNIST, the network is composed of 2 convolutive layers followed by a fully connected linear layer which outputs a vector of size $100$. For CIFAR10 and SVHN we use the same settings, but with 4 convolutive layers and the resulting final embedding, which size is $200$. Each head classifier is composed of 3 fully connected layers with a dropout function between consecutive layers, with drop probability set to $20\%$. In every network, we used ReLU as an activation function.

We experimented with different sizes for the external memories of ER, GEM, \changemarker{Replay}, and PRER, and for the latter, we also tested different ways of conditioning the generation of the embeddings. From the preliminary experiments, exposed in Section \ref{section:additional_results}, we take the best model and we average the score over $5$ experiments. In each of the $5$ experiments, we split the dataset randomly (balanced split based on labels, with proportion 80\% train and 20\% test set) and we change the initialization of the weights. A seed is assigned to each experiment, corresponding to the number of the experiment itself, to guarantee the same starting point for all the methods.

\subsection{Results}
\begin{table*}[t!]
\begin{center}
\caption{Average percentage on 5 runs, and the associated standard deviation, for BTW and Accuracy. All the results are calculated on the test sets. Best results within standard deviation are reported in \textbf{bold}.}
\label{table:results}
\vskip 0.15in
\begin{small}
\begin{sc}
\resizebox{0.9\linewidth}{!}{%
\centering
    \begin{tabular}{c|c|c|c|c|c|c|c|c|}
    \cline{2-9}
    \multirow{2}{*}{}                 & \multicolumn{2}{c|}{MNIST} & \multicolumn{2}{c|}{SVHN} & \multicolumn{2}{c|}{CIFAR10} & \multicolumn{2}{c|}{CIFAR100} \\ \cline{2-9} 
                                      & BWT                     & Accuracy                 &  BWT        & Accuracy    &  BWT        & Accuracy &  BWT        & Accuracy   \\ \hline
    \multicolumn{1}{|c|}{Naive}       &$-3.81_{\pm 1.40}$&$96.94_{\pm 2.76}$&$-4.80_{\pm 0.61}$&$94.12_{\pm 1.01}$&$-14.60_{\pm 1.01}$&$77.43_{\pm 0.99}$ &$-31.70_{\pm 1.21}$&$35.40_{\pm 1.46}$  \\ \hline
    \multicolumn{1}{|c|}{EWC \cite{Kirkpatrick2017overcoming}}         &$-0.28_{\pm 0.17}$&$99.64_{\pm 0.14}$&$-0.60_{\pm0.29}$&$95.34_{\pm 0.81}$&$-2.21_{\pm0.29}$&$80.88_{\pm0.81}$  &$-20.32_{\pm 0.98}$&$45.67_{\pm 1.12}$ \\ \hline
    \multicolumn{1}{|c|}{Replay \cite{chaudhry2019tiny}}   &$-3.81_{\pm 1.40}$&$91.09_{\pm 2.76}$&$-4.80_{\pm 0.61}$&$94.12_{\pm 1.01}$&$-17.54_{\pm 2.10}$&$85.14_{\pm 0.99}$ &$-20.85_{\pm 1.12}$&$49.47_{\pm 1.78}$ \\ \hline
    \multicolumn{1}{|c|}{GEM \cite{lopez2017gradient}}         &$-0.13_{\pm 0.01}$&$99.77_{\pm 0.02}$&$-0.90_{\pm0.12}$&$96.77_{\pm0.51}$&$-3.54_{\pm0.09}$&$86.81_{\pm 0.20}$ &$-25.40_{\pm 0.98}$&$40.77_{\pm 2.15}$\\ \hline
    \multicolumn{1}{|c|}{ER \cite{pomponi2020er}}          & $-0.11_{\pm 0.02}$&$99.78_{\pm 0.02}$&$\bm{-0.38_{\pm0.07}}$&$96.54_{\pm0.09}$&$-2.86_{\pm 0.46}$&$87.44_{\pm 0.36}$ &$-8.28_{\pm 0.76}$&$51.35_{\pm 1.06}$\\ \hline
    \multicolumn{1}{|c|}{GFR-IL \cite{liu2020generative}}          & $-0.10_{\pm 0.01}$&$99.72_{\pm 0.02}$&$\bm{-0.39_{\pm0.08}}$&$96.02_{\pm0.11}$&$-2.64_{\pm 0.32}$&$86.35_{\pm 0.36}$ &$-8.96_{\pm 0.81}$&$50.12_{\pm 0.68}$\\ \hline
    \multicolumn{1}{|c|}{SupSup \cite{wortsman2020supermasks}}   &-&$98.39_{\pm 0.15}$&-&$74.25_{\pm 0.85}$&-&$69.98_{\pm 0.07}$  &-&$44.08_{\pm 2.13}$ \\ \hline
    \multicolumn{1}{|c|}{PRER-R} &$\bm{-0.12_{\pm 0.06}}$&$\bm{99.73_{\pm 0.01}}$&$-0.50_{\pm 0.07}$&$96.45_{\pm 0.42}$&$-8.71_{\pm 0.12}$&$84.07_{\pm 0.18}$ &$-9.42_{\pm 0.57}$&$50.31_{\pm 0.78}$\\ \hline
    \multicolumn{1}{|c|}{PRER} &$\bm{-0.11_{\pm 0.05}}$&$\bm{99.41_{\pm 0.01}}$&$\bm{-0.38_{\pm 0.03}}$&$\bm{97.51_{\pm 0.21}}$&$\bm{-2.24_{\pm 0.09}}$&$\bm{88.25_{\pm 0.12}}$ &$-7.93_{\pm 0.72}$&$53.02_{\pm 1.02}$\\ \hline
    \end{tabular}%
    }
\end{sc}
\end{small}
\end{center}
\vskip -0.1in
\end{table*}
In this section, we expose the main results in terms of scores. In Table \ref{table:results} we summarize the results concerning the accuracy and BTW on the datasets exposed before.
First of all, we notice that in this multi-head scenario, the accuracy of the Naive method is only partially affected by the CF phenomenon, and some accuracy from the past tasks is preserved, depending on the difficulty of the dataset. The BTW values, however, are always significantly worse, highlighting the need for evaluating each method along multiple metrics.

Looking at EWC, we can see that it performs slightly better than the Naive approach. We observe that the scores are good for the first tasks, then, because the regularization penalty slows down the changing of the weights, the accuracy is negatively affected. With the growing complexity of the dataset this phenomenon becomes more visible: the accuracy is low but the remembering is aligned with the results obtained by better methods.

\changemarker{The others baseline methods perform better than EWC. As expected, ER is the best baseline method, as also reported in the original paper.}

The method we propose, PRER, achieves better or comparable results on all the benchmarks. Being based on ER, it is expected that the results are at least comparable to the ones achieved by ER. In our case, since we use a generative model instead of a fixed memory, the regularization process works better, leading to higher accuracy and lower forgetting. 

The results exposed are the ones obtained with the best hyper-parameters found doing a grid search, based on the results reported in the original papers. As we will see, when it is taken into account also the memory required by the method, PRER is the one which achieves the best results even with low memory usage. 
\begin{figure}[!t]
\vskip -0.2in
\centering
\begin{subfigure}{0.85\linewidth}
  \includegraphics[width=\linewidth]{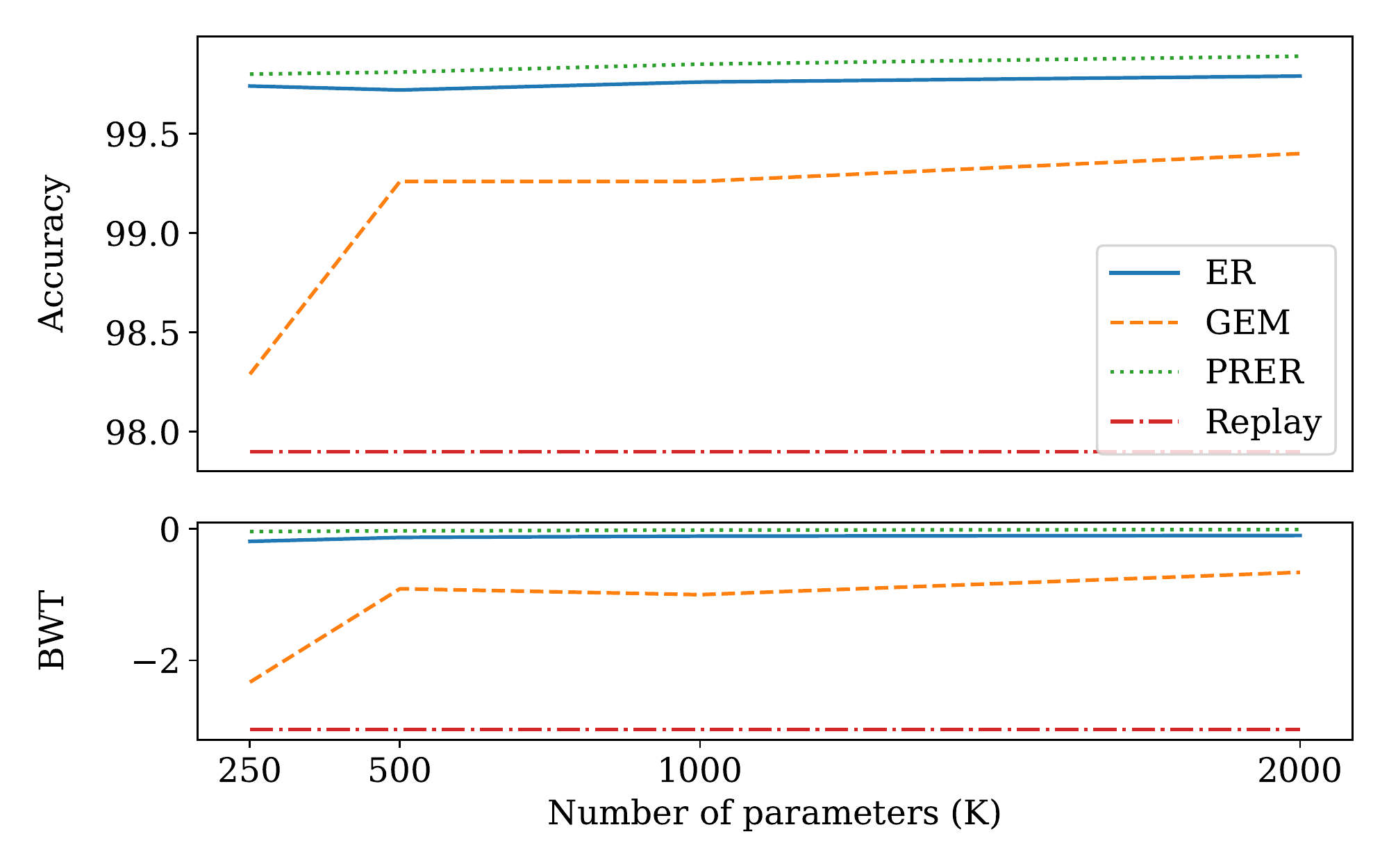}
  \caption{How the Accuracy and the BWT on MNIST change while varying the dimension of the memory.}
  \label{fig:ablation_memory_mnist}
\end{subfigure}
\begin{subfigure}{0.85\linewidth}
  \includegraphics[width=\linewidth]{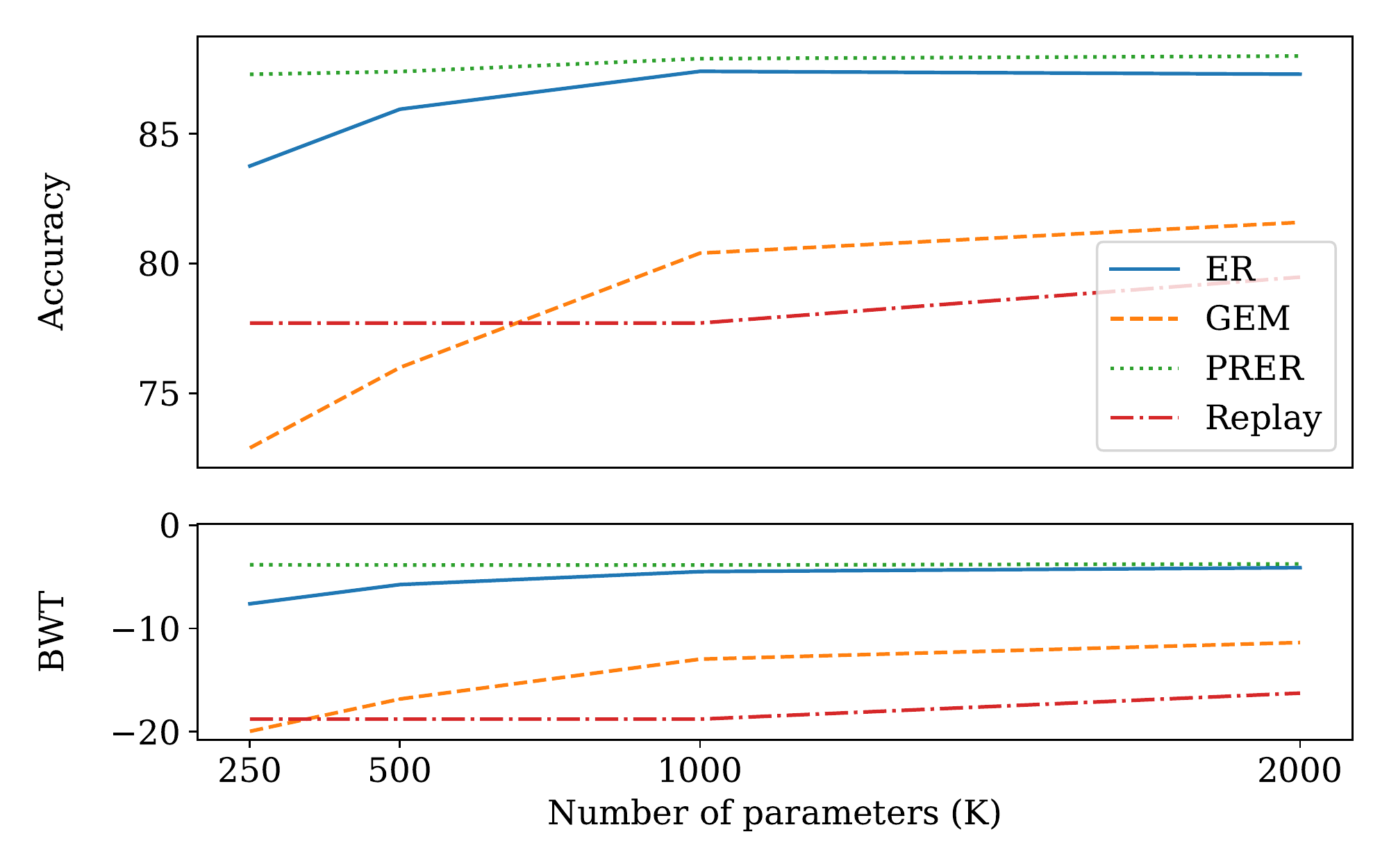}
  \caption{How the Accuracy and the BWT on CIFAR10 change while varying the dimension of the memory.}
  \label{fig:ablation_memory_cifar10}
\end{subfigure}
\caption{The images show how the dimension of the external memory affects the final results. For GEM and ER the parameters of the memory is the number of floats saved in memory, which is composed by images for GEM, with the addition of the embeddings vectors for ER. The memory required by PRER is the total number of trainable weights required by the encoder and the generative model.}
  \label{fig:ablation_memory}
  \vskip -0.2in
\end{figure}
\subsubsection{Hyperparameters}
\label{section:Hyperparameters}
To compare the size of the memories we approximate its dimension by counting the number of digits saved (the quantity of floats).

For \changemarker{ER, GEM, and Reaplay} we have one main parameter: the number of images to save in the external memory. The best results for ER and \changemarker{Replay} are achieved by setting the memory to 200 for all the datasets, resulting, approximately, in $676K$ parameters for MNIST and $3272K$ parameters for SVHN and CIFAR10. For GEM we set the number of saved images per task to $2000$, which approximately results in a memory of $5760K$ parameters for MNIST and $30720K$ parameters for CIFAR10 and SVHN. 
EWC has no memory to compare, and the only parameter is the importance of the past tasks (the value which weights the regularization penalty in the loss function), which is set to $10^3$ for each experiment.

PRER has many parameters and most of them are related to the generative model and its topology. As exposed before, our NF is composed of $L$ layers, and each one is made up of a number of blocks $B_l$; each block is composed of a permutation of the input, followed by a coupling layer and a batch norm layer to stabilize the training. The remaining parameters are the distance function $d(\cdot, \cdot)$, set to cosine distance in all experiments, and the regularization parameter $\beta$, fixed to $1$. As we will see, the dimension of the NF necessary to achieve good scores is contained. In fact, the best results are obtained using $1\text{-}3$ levels, with $5\text{-}10$ blocks for each level. In the end, the coupling layers take as input a vector of size $n$ and project it into a vector of size $2n$, before projecting it to the same input dimension $n$. 
The NFs which achieved the scores presented above have $250K$ parameters for MNIST and $500K$ for CIFAR10 and SVHN. Another architectural choice related to PRER is whether to condition or not the generation of the images; the results exposed are achieved when only the decoder is conditioned. In Section \ref{section:conditioning} the different ways to condition the models are exposed in detail. 

\subsection{Additional results}
\label{section:additional_results}
\subsubsection{Memory comparison}
\label{section:memory_comparison}
In this section, we study how the dimension of external memory influences the results. 
In order to compare the required memory, which we approximate as the number of floats saved, we need to define some quantities: $N$ is the number of trainable parameters of the encoder and the decoder, $M$ is the number of tasks, $IM$ is the dimension of an image, $EM$ is the dimension of the embeddings vector extracted by $E_c(\cdot)$, and $S$ is the number of samples saved in the external memory. Using these values, we can calculate the size of a memory, which depends on the method used, as follows: 

\begin{itemize}
    \item \changemarker{GEM-Replay: these methods save images} from past tasks, so we have that the memory size depends on the number of saved images, and it is given by the following formula: $M \times S \times IM$. 
    \item ER: this method, similarly to GEM, requires storing past images along with the embeddings vectors associated with them, extracted by the encoder $E_c(\cdot)$. The final memory dimension is calculated as $M \times S \times (IM + EM)$.
    \item PRER: it requires storing the decoder, along with the NF. The best way to count the number of parameters, which corresponds to the number of trainable parameters of the models, is empirical. 
    \end{itemize}
It is clear that the required memory depends mostly on the dimension of the images, with the addition of the embeddings size for ER. The memory required by our method depends on how deep and wide the NF is, in combination with the size of the embedding vectors, which is the input of our generative model. Also, we do not include EWC in the overall memory comparisons, since it does not use external memory. 

Fig \ref{fig:ablation_memory} shows the results obtained on MNIST and CIFAR10 while varying the memory budget. We decided different interesting values and managed to build the methods in order to match, approximately, the memory size with the budget (\emph{e.g.} to reach $250K$ parameters using GEM on MNIST, whose images have size $1\times28\times28$, we need to save $\approx 63$ images per task). \changemarker{We can see that GEM and Replay are the methods which suffer} more from the memory dimension since a lot of images are required to estimate correctly the direction of the gradients (the best results are obtained with $2000$ images for each task). Comparing ER with our proposal shows that both converge to the same score, but PRER requires less memory to reach it. This phenomenon is more evident when looking at the results obtained on CIFAR10. 

\begin{table}[t]
\caption{The table shows the results obtained using PRER while varying which network is conditioned and how it is done. Both of the networks, the decoder and the NF, can be conditioned using the categorical vector, one-hot, associated with a class.}
\label{table:ablation}
\centering
\resizebox{0.85\linewidth}{!}{%
\begin{tabular}{|c|c|c|c|c|}
\hline
\multirow{2}{*}{Autoencoder} & \multirow{2}{*}{Normalizing Flow} & \multirow{2}{*}{Dataset} & \multicolumn{2}{c|}{Results} \\ \cline{4-5} 
 & &  & \multicolumn{1}{c|}{BWT} & \multicolumn{1}{c|}{Accuracy} \\ \hline \hline
\multirow{2}{*}{Conditioned} & \multirow{2}{*}{Conditioned} & MNIST & $-0.34$ & $99.66$  \\ \cline{3-5} 
 &  & CIFAR10 & $-4.66$ & $87.22$ \\ \hline \hline
\multirow{2}{*}{Conditioned} & \multirow{2}{*}{-} & MNIST & $-0.07$ & $99.84$ \\ \cline{3-5} 
 &  & CIFAR10 & $-2.74$ & $88.33$ \\ \hline \hline
\multirow{2}{*}{-} & \multirow{2}{*}{Conditioned} & MNIST & $-0.07$ & $99.80$ \\ \cline{3-5} 
 &  & CIFAR10 & $-4.78$ & $86.63$ \\ \hline \hline
 \multirow{2}{*}{-} & \multirow{2}{*}{-} & MNIST & $-0.24$ & $99.70$ \\ \cline{3-5} 
 &  & CIFAR10 & $-4.09$ & $87.28$ \\ \hline
\end{tabular}%
}
\end{table}

\subsubsection{Conditioning techniques}
\label{section:conditioning}
In our experiments we tried different approaches to force the generative process to create images associated to a specific class.

In PRER we have two different networks that can be conditioned using a one-hot vector associated to a given label: the NF and the decoder. The key idea is that the models should be capable of generating meaningful embeddings, and thus images. It is not important if the models are or not conditioned, since in this scenario the labels are not used to regularize the training and will not be used after the generation.
The key idea is that, by conditioning the generation of the images, we can keep a proportionality between the classes in the generated images, by choosing which one to generate. 


Having two additional networks, we have four possible conditioning approaches: 

\begin{itemize}
    \item Both the networks are conditioned: in this case we do not rely on a single network to reconstruct the images but the NF is capable of recreating the important features associated to a class, and the decoder to reconstruct those features, helped by the class.   
    \item Only the NF is conditioned: in this case the NF, being conditioned, should focus more on the features which are related to the class.
    \item Only the decoder is conditioned: this case is the inverse of the one exposed before.
    \item Neither the decoder nor the NF are conditioned: this approach is possible because in a multi-head scenario the label is not needed to regularize the training.
\end{itemize}

In Table \ref{table:ablation} are shown the results associated with all the combinations exposed. 
We see that the cases in which the NF is conditioned give, in general, worse results. 
This happens because forcing the NF to learn the embeddings features associated to a given class leads it to place more mass on the features that the NF can relate to those classes, without taking into account that others features in the embeddings vector can be useful too. In this way, some features useful for classifying correctly the images could be missing, and also the reconstruction of the images can be worse for the same reason. 

The best model is the one in which only the reconstruction is conditioned. In this way the decoder, in order to minimize the reconstruction loss, forces the creation of embeddings that are more separable, and the NF is free to learn the distribution of the embeddings. For the same reason, the scenario in which both of the networks are conditioned leads to the worst results: being both conditioned, a mismatch between the important features selected by the decoder and the NF can arise.

\subsubsection{Quality of the generation}
\begin{figure*}[t]
\begin{subfigure}{\textwidth}
\begin{subfigure}{0.24\textwidth}
  \includegraphics[width=\linewidth]{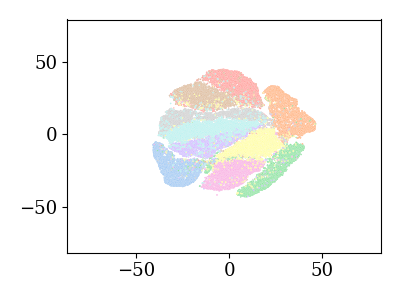}
\end{subfigure}\hfill
\begin{subfigure}{0.24\textwidth}
  \includegraphics[width=\linewidth]{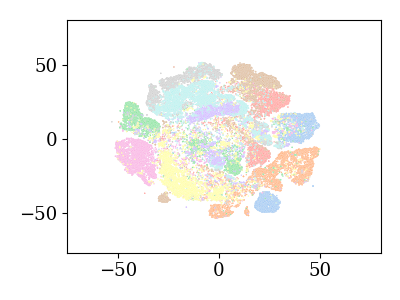}
\end{subfigure}\hfill
\begin{subfigure}{0.24\textwidth}%
  \includegraphics[width=\linewidth]{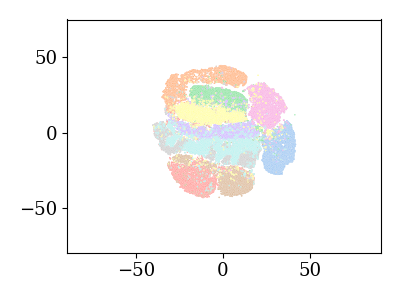}
\end{subfigure}\hfill
\begin{subfigure}{0.24\textwidth}%
  \includegraphics[width=\linewidth]{ab2_task4.png}
\end{subfigure}
\end{subfigure}
\vfill
\begin{subfigure}{\textwidth}
\begin{subfigure}{0.24\textwidth}
  \includegraphics[width=\linewidth]{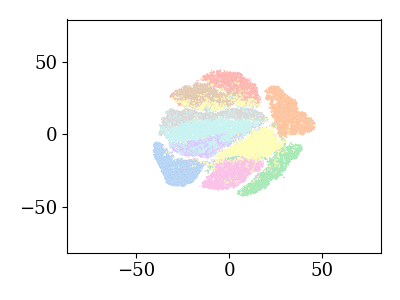}
\end{subfigure}\hfill
\begin{subfigure}{0.24\textwidth}
  \includegraphics[width=\linewidth]{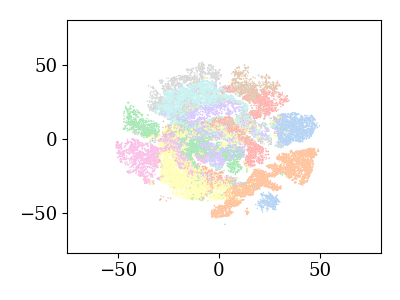}
\end{subfigure}\hfill
\begin{subfigure}{0.24\textwidth}%
  \includegraphics[width=\linewidth]{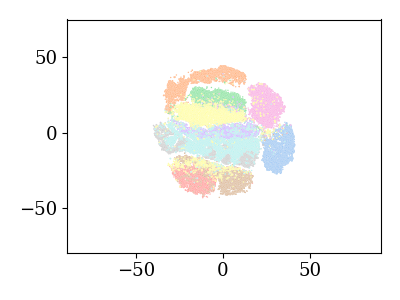}
\end{subfigure}\hfill
\begin{subfigure}{0.24\textwidth}%
  \includegraphics[width=\linewidth]{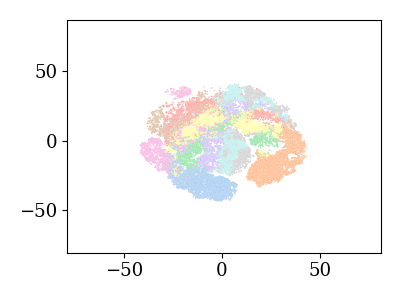}
\end{subfigure}
\end{subfigure}
\caption{The images show the embeddings spaces (using t-SNE \cite{maaten2008visualizing}) after the training on all the tasks from MNIST. The first row contains the embeddings produced by the autoencoders, while the second one the ones sampled from the generative models. Each column correspond to a different conditioning technique, respectively: 1) both of the networks are conditioned, 2) only the decoder is conditioned, 3) only the NF is conditioned, and 4) neither the decoder nor the NF are conditioned. We can notice that the generative model is capable of reproducing the embedding space produced by the autoencoder with high fidelity. Best viewed in color.}
\label{fig:embeddgins}
\end{figure*}
In this section we want to study the quality of the generated embeddings. Our regularization method does not requires that the generated images are realistic, but only that they contain all the information that can be useful to extract the embeddings vector used to regularize the training. For these reasons we evaluate the generation's quality by calculating how much the generated embeddings space is close to the real one and how much the generated embeddings vectors are preserved when reconstructing the images. 

In the first case, taken all the images with a given label $y^d$, we calculate all the embeddings associated to the images $E_r(\mathbf{x})$, producing the set $\mathbf{e}_{y^d} $. Then, we sample the same number of embeddings from the NF, by conditioning the reconstruction to $y^d$, producing the set of synthetic embeddings $\mathbf{e}_s$  (if the model is not conditioned, we classify the sampled embeddings using a Random Forest trained on $\mathbf{e}_{y^d} $). Given these two sets, the coverage metric is the Hausdorff distance between the real and the generated set, at the end of the task $t$: 

\begin{equation}
    \mathcal{D}_t(\mathbf{e}, \mathbf{e}_s) = \max 
    \Big\{
    \sup_{x\in \mathbf{e}} \inf_{y \in \mathbf{e}_s} d(x, y), 
    \sup_{y\in \mathbf{e}_s} \inf_{x \in \mathbf{e}} d(x, y)
    \Big\}
\end{equation}

\noindent where $d(\cdot, \cdot)$ is the Euclidean distance. We calculate this distance for each class $y^d$ encountered up to the task $t$, then we average the results.  

For evaluating how much information is preserved from the generation to the regularization, we sample $N$ tuples $\{ ( \hat{\mathbf{x}}_i, \hat{\mathbf{z}}_i) \}_{i=0}^{N}$ from the NF model trained on task $t$, as exposed in Section \ref{sec:regularization}, and calculate the quality as:

\begin{equation}
    \mathcal{Q}_t = \frac{1}{N} \sum_{i=1}^N s(E_c(\hat{\mathbf{x}}_i), \hat{\mathbf{z}}_i)
\end{equation}

\noindent where the similarity metric $d(\cdot, \cdot)$ is the inverse of distance used during the regularization process (cosine similarity in our case).

First of all, we want to show visually the coverage of the embeddings sets on MNIST. Fig. \ref{fig:embeddgins} contains $4$ columns, one for each conditioning technique, divided in the real embeddings space, top row, and the generated one, bottom row. It is interesting to see that the conditioning technique used changes not only the final scores, but also the final embedding space, because the generated embeddings vectors are also used to regularize the network; so, in base of the learned space by the NF, during the training the embeddings are moved towards a space in which the regularization distance is minimized. In fact, the best model, second column, is the one that has the most extended embeddings space, while the others tend to collapse the classes all together.

Table \ref{table:ablation_quality} shows the results obtained on MNIST and are related to the ones obtained in the ablation experiments exposed in Section \ref{section:conditioning}. We can see that not always the conditioning technique which achieves good scores is the one which produces the best quality score $\mathcal{Q}$, while the sets coverage remains ones of the best during all the training. The others results confirm what has been said before: the first model is the one which has more distance between the real embeddings set and the generated one due to the non-alignment of the embeddings, and, in general, the models in which the NF is conditioned are the ones which give worst results in terms of set coverage.

These results show that the quality of the generation can be related to the final scores, but should not be the only metric to guide the choice of the model to use in the final experiments.   
\begin{table}[t]
\caption{The table shows the quality results obtained on MNIST using PRER, while varying the conditioning technique. We used two quality measures: the first one is $\mathcal{D}$, which indicates how good is the coverage of the generated embeddings with respect to the real one produced by the encoder $E_c(\cdot)$, the second one, $\mathcal{Q}$, measures how much information from the generated embeddings is preserved when reconstructing the associated images.}
\label{table:ablation_quality}
\centering
\resizebox{\columnwidth}{!}{%
\begin{tabular}{|c|c|c|c|c|c|c|c|}
\hline
\multirow{2}{*}{Autoencoder} & \multirow{2}{*}{NF} & \multicolumn{5}{c|}{$\mathcal{D}_t$} & \multirow{2}{*}{$\mathcal{Q}_5$} \\ \cline{3-7}
 &  & $t=1$ & $t=2$ & $t=3$ & $t=4$ & $t=5$ &  \\ \hline
Conditioned & Conditioned & $5.30$ & $10.67$ & $21.32$ & $38.61$ & $47.02$ & $99.53$ \\ \hline
Conditioned & - & $6.02$ & $13.46$ & $24.24$ & $31.47$ & $38.91$ & $99.25$ \\ \hline
- & Conditioned & $6.13$ & $10.30$ & $21.04$ & $35.77$ & $40.49$ & $99.32$ \\ \hline
- & - & $4.27$ & $10.11$ & $17.66$ & $28.31$ & $34.41$ & $99.44$ \\ \hline
\end{tabular}%
}
\end{table}
\section{Conclusion}
In this paper, we introduced PRER (Pseudo-Rehearsal Embedding Regularization), a pseudo-rehearsal method that, by working on the embedding space, is able to generate past embeddings and use them to protect past information while learning new tasks. This approach is a different point of view on the pseudo-rehearsal methods, which usually work on the input space. By working on a lower complexity space, the required dimension of the generative model is reduced. Once combined with a decoder, the generative model can be used to generate images associated with past tasks and to constrain the model by regularizing the embeddings' deviation. We believe that this set of methods can be further investigated, leading to a different view of the pseudo-rehearsal approaches, which, right now, are feasible only for low-complexity datasets. We leave an investigation on how the PRER method scales to more complex datasets for future work. \changemarker{Moreover, it is worth investigating how to extend PRER to work on class incremental scenarios (by, for example, creating an external memory associated with past tasks, that can be used to regularize the generative approach of the method)}.  
    \bibliographystyle{elsarticle-num}
\balance
\bibliography{main}
\end{document}